\pdfoutput=1

\documentclass[11pt]{article}

\usepackage[final]{acl}

\usepackage{booktabs}
\usepackage{longtable}
\usepackage{times}  
\usepackage{latexsym}
\usepackage{makecell}
\usepackage{amsfonts}
\usepackage{amsmath}
\usepackage{booktabs}
\usepackage{enumitem}
\usepackage[utf8]{inputenc}
\usepackage{multirow}
\usepackage{threeparttable}
\usepackage{float}
\usepackage{array}
\usepackage{tikz}
\usetikzlibrary{positioning}
\usepackage{varwidth}
\usepackage{graphicx}

\usepackage[T1]{fontenc}

\usepackage[utf8]{inputenc}

\usepackage{microtype}

\usepackage{inconsolata}

\usepackage{graphicx}

\usepackage{xtab}
\title{Evaluating Uncertainty Quantification Methods \\ in Argumentative Large Language Models}

\author{
\\
  \textbf{Kevin Zhou}\textsuperscript{1}\quad
  \textbf{Adam Dejl}\textsuperscript{1}\quad
  \textbf{Gabriel Freedman}\textsuperscript{1}\\
  \textbf{Lihu Chen}\textsuperscript{1}\quad
  \textbf{Antonio Rago}\textsuperscript{1,2}\quad
  \textbf{Francesca Toni}\textsuperscript{1}
\\
  \textsuperscript{1}Imperial College London \quad
  \textsuperscript{2}King's College London
\\
 \texttt{\{kevin.zhou24,adam.dejl18,g.freedman22,lihu.chen,ft\}@imperial.ac.uk} \\ \texttt{antonio.rago@kcl.ac.uk}   
}

\begin{document}
\maketitle
\begin{abstract}
Research in uncertainty quantification (UQ) for large language models (LLMs) is 
increasingly important towards guaranteeing the reliability of this groundbreaking technology. 
We explore the integration of LLM UQ methods in argumentative LLMs (ArgLLMs), an explainable LLM framework for decision-making based on computational argumentation in which UQ plays a critical role. We conduct experiments to evaluate ArgLLMs' performance on claim verification tasks when using different LLM UQ methods, inherently performing an assessment of the UQ methods' effectiveness. Moreover, the experimental procedure itself is a novel way of evaluating the effectiveness of UQ methods, especially when intricate and potentially contentious statements are present. Our results demonstrate that, despite its simplicity, direct prompting is an effective UQ strategy in ArgLLMs, outperforming considerably more complex approaches. 
\end{abstract}

\section{Introduction}
Large language models (LLMs) have demonstrated impressive capabilities across a range of tasks, such as coding, reasoning, and speech recognition \cite{openai2024gpt4technicalreport, grattafiori2024llama3herdmodels}. However, they also often generate hallucinated answers \citep{sahoo2024}, with no clear indication of the uncertainty which caused them. Still, many users are prone to blindly trusting LLMs' responses \cite{klingbeil2024108352}, which is especially risky in areas such as healthcare where LLMs are being applied \citep{he2025survey}. In these settings, the ability to reliably retrieve an LLM's uncertainty would be immensely impactful, highlighting the importance of uncertainty quantification (UQ) research in the development of trustworthy AI systems.

Recent research has shown that LLMs exhibit strong performance in automated decision-making
~\cite{ouyang-li-2023-autoplan}, but that they also face challenges such as the inability to faithfully explain their decisions \cite{turpin-2023-unfaithful-cot, chen-2025-unfaithful-reasoning-models} and reliably correct mistakes based on user feedback \cite{freedman2024}. Towards addressing these challenges, \citet{freedman2024} introduce argumentative LLMs (ArgLLMs), which leverage computational argumentation to improve explainability and contestability for decision-making tasks such as claim verification.

\begin{figure}[t]
    \centering
    \includegraphics[width=0.50\textwidth]{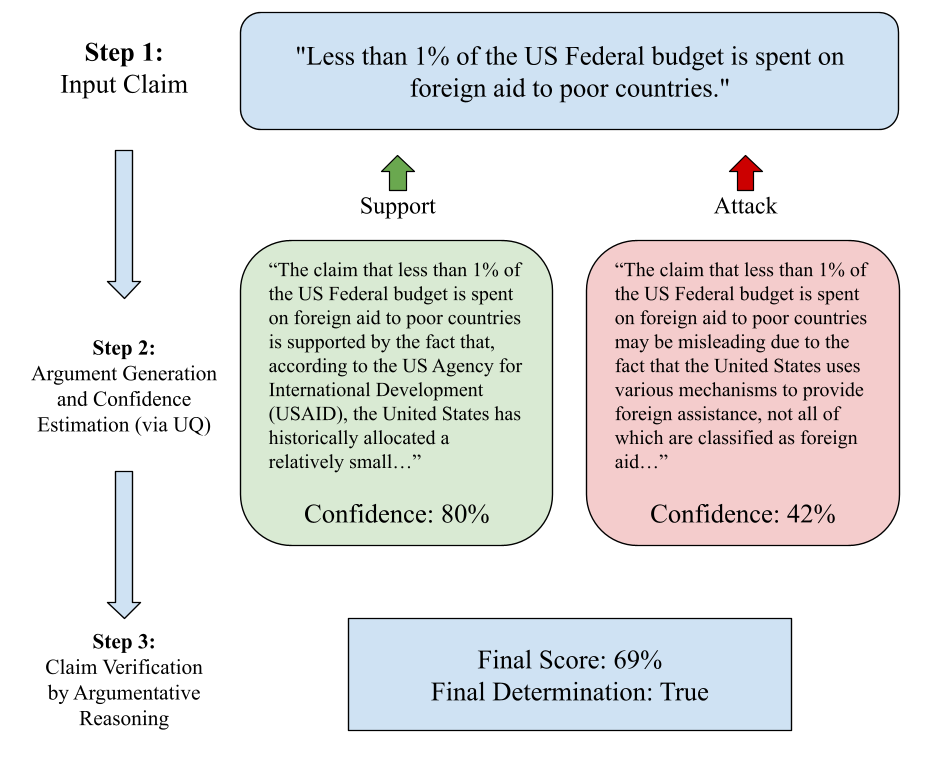}
    \caption{Example 
    of argumentative LLMs,
    where UQ plays a crucial role in estimating confidence in the generated arguments and thus in the claim verification itself.
    Here, the input claim is derived 
    from the TruthfulQA dataset \cite{lin2022truthfulqameasuringmodelsmimic}, arguments are generated by Llama 3.1 \cite{grattafiori2024llama3herdmodels}, and a default base score 0.50 is used for the input claim.}
    \label{fig:arg-llm}
\end{figure}

For a given statement, ArgLLMs generate an argumentation framework of supporting and attacking arguments, quantify argument uncertainty, and aggregate these components using gradual semantics \cite{baroni2019}, a form of formal reasoning, to determine whether a statement is true 
(see Figure \ref{fig:arg-llm} for an example of an ArgLLM evaluating a claim).
ArgLLMs demonstrate comparable performance in 
claim verification 
against prompting methods such as direct questioning and chain-of-thought \cite{wei2022cot} while also providing faithful explanations through the argumentative structure. 
 

It can be seen from Figure \ref{fig:arg-llm} that in ArgLLMs, the estimated confidence in the generated arguments, and thus the chosen UQ method, is pivotal in the claim verification task itself. For example, if (unlike in Figure \ref{fig:arg-llm}) the confidence in the attacking argument(s) was stronger than that in the supporting argument(s), we would expect the final determination to be false.\footnote{Note that the argumentative reasoning generated by ArgLLMs can be significantly more complex and deep than the example shown in Figure \ref{fig:arg-llm}.} Thus, ArgLLMs are an appealing evaluation setting for UQ methods. 

Since the confidence scores from the used UQ method directly feed into the final predictions, ArgLLMs enable evaluation of these scores without requiring access to the ground-truth labels for the arguments, as they only require the label for the top-level claim. For instance, arguments in favor of an incorrect prediction can generally be expected to be less convincing, and should thus be less certain. In this setting, UQ methods can be seen as a proxy for downstream factuality and output reliability, which is a common use-case. Additionally, ArgLLMs involve long and potentially contentious arguments, and the generation of supporting and attacking arguments for each claim ensures a strong diversity of statements to evaluate, making for a challenging, realistic, and wide-ranging setting. 

Various UQ methods exist in the literature, from directly prompting the model for an uncertainty estimate (also known as verbalized UQ), which have proven effective and at times well-calibrated \cite{yang2024verbalizedconfidencescoresllms, tian-etal-2023-just}, to more complex methods involving token logits or semantic similarity \cite{geng-etal-2024-survey}. It is thus unclear which UQ method would be ideal in the context of ArgLLMs, which use direct prompting in \citet{freedman2024}. This is especially true since verbalized UQ can sometimes have issues arising from LLMs being overconfident and biased when evaluating their own answers \cite{sun2025largelanguagemodelsoverconfident}. In this paper, we aim to shed light on this problem by experimenting with a set of LLM UQ methods and examining their effect on the performance of ArgLLMs in the claim verification task. Concretely, our contributions are as follows:
\begin{itemize}[nosep]
    \item We integrate several LLM UQ methods that have performed well in long-form generation benchmarks \cite{vashurin2024,zhang2024luq} into the explainable framework of ArgLLMs, presenting a novel method for evaluating their effectiveness via the resulting accuracy in downstream claim verification. 
    \item We conduct experiments with three claim verification datasets, three LLMs, and four ArgLLM settings, resulting in $36$ different configurations where we evaluate three LLM UQ methods as well as the direct prompting baseline. The results show direct prompting outperforms the other UQ methods. 
\end{itemize}
\section{Background and Preliminaries}

\subsection{Uncertainty Quantification}
In addition to direct prompting, we use three 
LLM UQ methods: Semantic Entropy \cite{kuhn2023semantic}, Eccentricity \cite{lin2024eccentricity}, and LUQ \cite{zhang2024luq}. For these additional methods, we focus on the version of each method implemented in the MIT-licensed LM-Polygraph library \cite{fadeeva2023lmpolygraph}
. Further details of the methods 
are given in Appendix \ref{sec:appendix-uq}.

{\bf Direct Prompting}
involves the model directly providing a confidence score for the inputted text. The prompt used to obtain confidence scores for direct prompting is given in Appendix \ref{sec:appendix-prompts}.

{\bf Semantic Entropy}
Each time the model is prompted
, it generates (a hyperparameter) $M$ different samples. Then, samples with similar meaning are clustered 
and the entropy over the meaning-distribution determined by 
the clustering is computed using token logits. 

{\bf Eccentricity}
has multiple variations: in this paper we use the natural language inference (NLI) Entailment version. 
After generating $M$ samples for an input, an NLI classifier model computes entailment logits between the generations, uses them to calculate similarity scores, and then constructs a graph Laplacian with the similarity scores as edge weights. The uncertainty is then computed as the average distance from the center of the eigenvectors, with the intuition being that a lower uncertainty would result in more similar samples and thus a lower average distance. 


{\bf LUQ} (Long-text Uncertainty Quantification) also involves multiple generated samples. In \citet{zhang2024luq}, the generations are split into component sentences, but LM-Polygraph considers a simplified version which leaves the generations in their full form. An NLI model is used to obtain logit values of ``entailment'' and ``contradiction'' between each generated response, and then the uncertainty is computed as a function of both the ``entailment'' and ``contradiction'' logits.

Similar to Eccentricity and LUQ, other methods in the literature also make use of entailment and NLI-based scoring, highlighting the capabilities of NLI scores in uncertainty estimation processes. Examples of such methods include SelfCheckGPT \cite{manakul-etal-2023-selfcheckgpt} and HaLoCheck \cite{elaraby2023halo}, which are designed primarily for the task of hallucination detection. 

\subsection{Argumentative LLMs}
ArgLLMs 
construct 
quantitative bipolar argumentation frameworks (QBAFs), which consist of arguments connected through support and attack relations where each argument also has a base score representing its intrinsic strength \cite{baroni2019}. In the context of ArgLLMs, the supporting and attacking arguments have their base score set as the confidence score outputted by the UQ method.\footnote{In this work, the confidence score is considered the antonym of uncertainty.} QBAFs can then be evaluated via a gradual semantics \cite{baroni2019}, which determines the final strength of each argument, taking into account its intrinsic strength as well as the strengths of its attackers and supporters. 
In ArgLLMs, each 
input claim can have supporting and attacking arguments generated for it, and each of those arguments can have its own supports and attacks \cite{freedman2024}. From the scores and connections of these QBAF components, the DF-QuAD \cite{rago2016} gradual semantics is used to compute the original claim's strength. 
If the final score of the claim is greater than $0.5$, it is predicted to be true; otherwise it is predicted to be false. 

The structure of ArgLLMs provides a unique and also realistic setting for LLM UQ method evaluation where the confidence scores of generated arguments are integral to downstream claim verification, which to the best of our knowledge has not yet been studied in the literature. 

\section{Experiments}
We evaluate the performance of ArgLLMs on the claim verification task when using different LLM UQ methods as the uncertainty estimator for the arguments. 
For the 
experiments, we build upon the publicly released code provided by \citet{freedman2024}.\footnote{Our 
code is available at \url{https://github.com/CLArg-group/argumentative-llms-uq}. All experiments are 
run on a system with RTX 4090 24GB GPUs with the seed set to 42 for reproducibility.} We also employ the same prompts as \citet{freedman2024} except for a minor change to argument generation (see 
Appendix \ref{sec:appendix-prompts}).
\subsection{UQ Integration}
We use the LM-Polygraph implementations for Semantic Entropy, Eccentricity, and a simplified version of LUQ without sentence splitting, including LM-Polygraph's default value of $10$ for the number of samples generated per input. The performance of ArgLLMs with these methods is compared with the baseline UQ method of direct prompting. 
\subsection{Datasets}
We use the TruthfulClaim, StrategyClaim and MedClaim datasets \cite{freedman2024}, which are tailored versions of 
TruthfulQA \cite{lin2022truthfulqameasuringmodelsmimic}, StrategyQA \cite{geva2021didaristotleuselaptop}, and MedQA \cite{jin2021diseasedoespatienthave}
. 
Details for each dataset are discussed further 
in 
Appendix \ref{sec:appendix-datasets}.

An important consideration is the risk of data contamination, especially since these claim-based datasets are derived from popular QA datasets. We believe that in this case, the associated risk is mitigated by the nature of ArgLLMs. Specifically, using ArgLLMs substantially changes the original task and data distribution, as the model is generating attacking and supporting arguments and providing subsequent confidence scores for these arguments rather than directly answering the questions from the original QA datasets.

\begin{table*}[t]
    \centering
    \scriptsize
    \begin{tabular}{c|c|cccc|cccc|cccc}
        \multirow{3}{*}{\textbf{Model}} & 
        \multirow{3}{*}{\textbf{UQ Method}} & 
        \multicolumn{4}{c}{\textbf{TruthfulClaim}} &
        \multicolumn{4}{c}{\textbf{StrategyClaim}} &
        \multicolumn{4}{c}{\textbf{MedClaim}} \\
         &
         &
        \multicolumn{2}{c}{\textbf{D=1}} &  
        \multicolumn{2}{c}{\textbf{D=2}} &  
        \multicolumn{2}{c}{\textbf{D=1}} &  
        \multicolumn{2}{c}{\textbf{D=2}} &  
        \multicolumn{2}{c}{\textbf{D=1}} &  
        \multicolumn{2}{c}{\textbf{D=2}} \\
         &
         &
        \!\!\textbf{0.5 BS}\!\! &  
        \!\!\textbf{Est. BS}\!\! &
        \!\!\textbf{0.5 BS}\!\! &  
        \!\!\textbf{Est. BS}\!\! & 
        \!\!\textbf{0.5 BS}\!\! &  
        \!\!\textbf{Est. BS}\!\! &
        \!\!\textbf{0.5 BS}\!\! &  
        \!\!\textbf{Est. BS}\!\! & 
        \!\!\textbf{0.5 BS}\!\! &  
        \!\!\textbf{Est. BS}\!\! &
        \!\!\textbf{0.5 BS}\!\! &  
        \!\!\textbf{Est. BS}\!\! \\ 
        \hline
        \multirow{4}{*}{\makecell{Llama 3.1}} & 
        Direct Prompting & 
        \textbf{0.626} & 
        \underline{0.664} & 
        \textbf{0.652} & 
        \textbf{0.658} & 
        \underline{0.594} & 
        \underline{0.600} & 
        \underline{0.592} & 
        \underline{0.590} & 
        \textbf{0.594} & 
        \textbf{0.606} & 
        \textbf{0.574} & 
        \underline{0.604}\\
         &
        Semantic Entropy & 
        \underline{0.604} & 
        \underline{0.650} & 
        0.592 & 
        \underline{0.650} & 
        \underline{0.574} & 
        \underline{0.592} & 
        0.552 & 
        \underline{0.588} & 
        \underline{0.548} & 
        \underline{0.596} & 
        \underline{0.556} & 
        0.566\\
         & 
        Eccentricity & 
        0.514 & 
        0.606 & 
        0.508 & 
        \underline{0.638} & 
        0.530 & 
        0.566 & 
        0.534 & 
        0.566 & 
        0.484 & 
        0.518 & 
        0.490 & 
        0.548\\ 
         & 
        LUQ & 
        0.556 & 
        \textbf{0.668} & 
        0.552 & 
        \underline{0.654} & 
        \textbf{0.622} & 
        \textbf{0.614} & 
        \textbf{0.614} & 
        \textbf{0.600} & 
        \underline{0.578} & 
        \underline{0.594} & 
        \textbf{0.574} & 
        \textbf{0.608}\\
        \hline
        \multirow{4}{*}{\makecell{Gemma 2}} & 
        Direct Prompting & 
        \textbf{0.682} & 
        \underline{0.732} & 
        \textbf{0.674} & 
        \underline{0.732} & 
        \textbf{0.656} & 
        \textbf{0.708} & 
        \textbf{0.652} & 
        \textbf{0.702} & 
        \textbf{0.596} & 
        \underline{0.578} & 
        \textbf{0.576} & 
        \underline{0.582}\\
         & 
        Semantic Entropy & 
        0.516 & 
        \textbf{0.746} & 
        0.558 & 
        \underline{0.734} & 
        0.466 & 
        0.666 & 
        0.490 & 
        \underline{0.696} & 
        \underline{0.518} & 
        \underline{0.580} & 
        \underline{0.546} & 
        \underline{0.578}\\ 
         & 
        Eccentricity & 
        0.504 & 
        \underline{0.714} & 
        0.500 & 
        \textbf{0.740} & 
        0.464 & 
        0.634 & 
        0.450 & 
        \underline{0.676} & 
        \underline{0.534} & 
        \underline{0.580} & 
        0.496 & 
        \underline{0.582} \\
         &
        LUQ & 
        0.560 & 
        \underline{0.714} & 
        0.584 & 
        \underline{0.712} & 
        0.526 & 
        \underline{0.672} & 
        0.538 & 
        \underline{0.686} & 
        \underline{0.566} & 
        \textbf{0.590} & 
        \underline{0.542} & 
        \textbf{0.584}\\
        \hline
        \!\!\multirow{4}{*}{\makecell{GPT-4o-mini}}\!\! & 
        Direct Prompting  & 
        \textbf{0.748} & 
        \textbf{0.816} & 
        \textbf{0.764} & 
        \textbf{0.822} & 
        \textbf{0.646} & 
        \textbf{0.742} & 
        \textbf{0.690} & 
        \underline{0.736} & 
        \textbf{0.638} & 
        \textbf{0.718} & 
        \textbf{0.644} & 
        \textbf{0.710}\\
         & 
        Semantic Entropy & 
        N/A & 
        N/A & 
        N/A & 
        N/A & 
        N/A & 
        N/A & 
        N/A & 
        N/A & 
        N/A & 
        N/A & 
        N/A & 
        N/A \\
         &
        Eccentricity & 
        0.512 & 
        0.722 & 
        0.496 & 
        0.760 & 
        0.548 & 
        0.680 & 
        0.534 & 
        \underline{0.724} & 
        0.528 & 
        0.656 & 
        0.516 & 
        \underline{0.686} \\
         &
        LUQ & 
        0.610 & 
        0.780 & 
        0.618 & 
        \underline{0.796} & 
        \underline{0.610} & 
        \underline{0.722} & 
        0.632 & 
        \textbf{0.742} & 
        0.546 & 
        0.662 & 
        0.510 & 
        \underline{0.704}\\
        \hline
    \end{tabular}
    \caption{Accuracy ($\uparrow$, best in bold) of ArgLLMs in all experiments. 
    Values other than the best that are not statistically significantly worse than the best accuracy are underlined. Semantic Entropy results for GPT-4o-mini are marked as N/A for the reasons discussed in Section~\ref{sec:models}. In  the ``0.5 BS'' setting, the claim's base score is set at 0.5, while in ``Est. BS'' it is estimated through prompting.}
    \label{tab:results-table}
\end{table*}

\subsection{Models}
\label{sec:models}
The LLMs we use are Google's gemma-2-9b-it \cite{gemmateam2024gemma}, Meta's Llama-3.1-8B \cite{grattafiori2024llama3herdmodels}, and OpenAI's GPT-4o-mini \cite{gpt-4o-mini}. We choose these models because they have demonstrated strong performance on model benchmarks and fit within our compute resources. The gemma-2-9b-it and Llama-3.1-8B models are open-source\footnote{We 
adopt a broad notion of 
``open-source", not necessarily implying 
licenses approved by the Open Source Initiative.}, while GPT-4o-mini is closed-source. Notably, since Semantic Entropy requires direct access to model logits in its computations, the LM-Polygraph implementation is not compatible with GPT-4o-mini. All open-source models are quantized to 4 bits to lower the computational cost \cite{dettmers2023}.

\subsection{Experimental Procedure}
\label{sec:procedure}
Each experiment is defined by the dataset, LLM, UQ method, the method of determining the claim's base score, and the depth. A depth of 1 (i.e. D=1) means that for each claim, one supporter and one attacker will be generated, and a depth of 2 (i.e. D=2) means that each of those arguments will also have a supporter and attacker. For the claim's base score, we adopt the two methods from \citet{freedman2024} of either setting it to $0.5$ or estimating it by prompting the LLM. We are not able to use the other LLM UQ methods with these claims because they are pre-existing text from the datasets and are not generated by the LLM itself. 

Thus, for each claim, the LLM generates supporting and attacking argument(s), and uses the UQ method to obtain confidence scores for each argument. Importantly, the UQ methods besides the direct prompting baseline produce a raw score that is not necessarily in $[0, 1]$, which ArgLLMs require. To address this, we employ binned normalization by grouping the outputs into $20$ quantile bins linearly mapped to a score in $[0,1]$, which is more robust to skewed distributions than a strict linear normalization. The DF-QuAD semantics is then used to compute a final confidence measure, which determines the prediction. We adopt the accuracy of ArgLLMs on claim verification as the primary downstream performance metric to observe the impact of integrating the different UQ methods. All experiments use $500$ data samples, which are identical to those used by \citet{freedman2024}.

\begin{table}[h]
\small
\centering
\setlength{\tabcolsep}{2.5pt}
\begin{tabular}{lcc}
\toprule
\textbf{UQ Method} & \textbf{Best} & \textbf{\makecell{Not Significantly \\ Worse than Best}} \\
\midrule
Direct Prompting  & 25 (0.69) & 11 (0.31)\\
Semantic Entropy            &  1 (0.04) & 15 (0.63) \\
Eccentricity                &  1 (0.03) &  8 (0.22) \\
LUQ                         & 10 (0.28) &  13 (0.36) \\
\bottomrule
\end{tabular}
\caption{Summary of the accuracy results from Table \ref{tab:results-table}, counting the number of experiments in which each method performed best or did not perform statistically significantly worse than the best method. The values in the parentheses are the counts divided by the number of experiments the UQ method is used in.  The ``Best'' column counts add up to $37$ since LUQ and Direct Prompting tied for best with MedClaim, Llama 3.1, 0.5 BS, and D=2.}
\label{tab:uq-summary}
\end{table}

\section{Results and Discussion}
\label{sec:results}
Table \ref{tab:results-table} presents the accuracy results from all experiments, indicating that direct prompting clearly achieves the best performance. 
Table \ref{tab:uq-summary} shows that direct prompting is either the best UQ method or not statistically significantly worse than the best in all $36$ configurations, and its $25$ instances of being the best UQ method are by far the most. An important caveat for our results and subsequent conclusions is that the models used are relatively small compared with the leading LLMs, and each experiment is only run once. 

In some cases, the advantage of direct prompting was substantial, such as in the StrategyClaim Gemma-2 0.5 BS (D=1) setting where it results in a $0.130$ higher accuracy than the next best method. In contrast, the most direct prompting is ever outperformed is $0.028$ by LUQ in the StrategyClaim, Llama 3.1, 0.5 BS, and D=1 setting. 

We assessed the statistical significance of the accuracy results by conducting bootstrap tests with $5000$ resamples to obtain $95\%$ confidence intervals for the pairwise accuracy differences between the methods. These confidence intervals are used to determine the statistical significance of the best performances per configuration in Table \ref{tab:results-table}. Furthermore, across all $180$ confidence intervals comparing UQ method performances, $74$ indicate statistically significant differences. Of these, $44$ involve direct prompting having a statistically significant advantage, followed by $24$ for LUQ, $6$ for Semantic Entropy, and $0$ for Eccentricity. The table of confidence intervals and further details are included in Appendix \ref{sec:appendix-results}.

In addition to accuracy, we also measured the Brier scores for all experiments, which computes the mean squared difference between the predicted probability and the true label. In our case, the final ArgLLM score is used as the predicted probability, and the label is 1 if the topic claim is true and 0 if it is false. The full table of Brier scores is presented in Appendix \ref{sec:appendix-results}. In summary, direct prompting scored the best in 18 instances, the most of any method, followed by LUQ with 9, Semantic Entropy with 7, and Eccentricity with 2.

Overall, these results support the notion
that verbalized confidence scores from direct prompting can be well-calibrated \cite{tian-etal-2023-just} and represent the model's internal knowledge well with effective prompting \cite{yang2024verbalizedconfidencescoresllms}. 

Moreover, direct prompting likely outperforms the other methods due to the nature of long-form contentious generations in argumentation. Sampling-based methods such as Semantic Entropy require the capture of semantic consistency among multiple arguments, which can lead to degraded performance as the length of texts grows. As shown in Figure~\ref{fig:arg-llm},  ArgLLMs have long generations which can sometimes be contentious, unlike more definitive true or false statements. In this situation, direct prompting is often better suited to estimate a reasonable uncertainty level based on the LLM's self-knowledge.
Also, it does not rely on an additional normalization step to map the UQ outputs to suitable confidence scores, 
which can be prone to noise and introduces further estimation compounded with the existing estimation task.  


On top of its superior performance, the advantage of direct prompting is further amplified by its lower resource requirements. Many of the other high-performing LLM UQ methods, including all three tested in this paper, require a separate NLI model and multiple generations per instance, increasing both  memory and time complexity
.

While 
not outperforming direct prompting overall, LUQ's relatively strong performance and high frequency of performing the best are intriguing. For example, when using Llama 3.1 with StrategyClaim, LUQ is the best in all four settings, illustrating the potential for a UQ method to perform the best in a specific setup. In total, as seen in Table \ref{tab:uq-summary}, LUQ performs the best $10$ times, which is much greater than the $1$ time for each of Semantic Entropy and Eccentricity. However, it is worth noting that Semantic Entropy often closely trails behind the best performing method with no statistically significant difference, even if it is rarely the best method. 

LUQ's capabilities in these experiments are also consistent with its strong performance in factuality tasks \cite{zhang2024luq}. While sentence splitting is not included in the LUQ implementation we use, computing uncertainty more directly from the entailment and contradiction logits between responses is another potential advantage which could have contributed to the stronger performance. All in all, the experiments demonstrate that ArgLLMs offer a compelling benchmark for LLM UQ methods. The QBAF structure of supporting and attacking arguments poses challenges distinct from those in existing benchmarks. At the same time, certain features that enhance performance on other tasks 
can also improve ArgLLM claim verification performance as in the case of LUQ, lending further credibility to ArgLLMs as an evaluation environment.
\section{Conclusion}
Our work integrates commonly used and high-performing LLM 
UQ methods into 
ArgLLMs and assesses their performance 
on claim verification. We find that
direct prompting leads to notably better performance 
than the other UQ methods. Among the latter, the LM-Polygraph implementation of LUQ performs better than Semantic Entropy and Eccentricity, reflecting LUQ's advantages seen in other tasks.
Overall, the experiments affirm the role of verbalized confidence prompting in eliciting confidence scores in ArgLLMs and suggest that prompt-based methods offer benefits for LLM UQ with long-form and potentially contentious statements. Our research also presents and highlights the value of evaluating LLM UQ methods in argumentative settings and faithfully explainable frameworks such as ArgLLMs. 

\newpage
\section{Limitations}
We now discuss some limitations of our work, particularly with regard to the experiments:
\begin{itemize}[nosep]
    \item We did not conduct multiple runs for each configuration due to the high computational and time cost of each run, which limits the robustness of statistical analysis on the experiments.
    \item We had to choose models which together with the entire experimental pipeline could fit within our compute resources. This limited the possible size of LLMs in our experiments.
    \item As discussed in Section \ref{sec:procedure} and Section \ref{sec:results}, the use of binned normalization in the UQ process is a limitation and likely negatively impacts the calibration and performance of the additional UQ methods. Future work could benefit from a more robust and separately evaluated calibration procedure.
    \item We only evaluate the task of claim verification with true or false as the possible labels in this paper; future experiments with other types of datasets and tasks would further enrich our understanding of LLM UQ integration with ArgLLMs.
\end{itemize}



\section{Ethical Considerations}
One potential risk of UQ with ArgLLMs in general is that malicious actors could theoretically devise a bad-faith UQ method to output confidence scores in line with an agenda, and then integrate it into the background of ArgLLMs and present the ArgLLM outputs for means of persuasion or demonstration. As a result, it is paramount that any user presenting ArgLLM outputs is also transparent and truthful about the UQ method used. Additionally, some sample claims in the datasets may contain untrue stereotypes or beliefs. We do not endorse any of the statements or opinions in the datasets, and their only purpose in the experiments would be to serve as sample claims that the ArgLLM evaluates.
\bibliography{acl_latex_camera_ready}
\newpage
\appendix

\section{Uncertainty Quantification Methods}
\label{sec:appendix-uq}
\paragraph{Semantic Entropy}
The clustering in Semantic Entropy is performed based on the concept of bi-directional entailment; given two inputs, a natural language inference (NLI) model such as DeBERTa-large \cite{he2021} is used to determine if one entails the other and vice versa, and the generations are clustered together if both directions are true.
The likelihoods of each sample are summed together within a cluster using the token logits, and then the entropy is computed over the meaning-distribution to determine the semantic entropy for the input text. 
Formally, \citet{kuhn2023semantic} express the semantic entropy as: 
\begin{multline}
    SE(x) = -\sum_{c}p(c|x)\log p(c|x) \\ = -\sum_{c}((\sum_{s\in{c}}p(s|x))\log[\sum_{s\in{c}}p(s|x)])
\end{multline}
where $x$ is the input text, $c$ represents a semantic equivalence class, and $s$ is a sequence. However, in practice, only the distribution generated by the model is accessible for the calculations. Thus, the semantic entropy is estimated through Monte Carlo integration as such: 
\begin{equation}
    SE(x) \approx -|C|^{-1}\sum^{|C|}_{i=1}\log p(C_i|x)
\end{equation}
where $C$ is the set of semantic equivalence classes induced by the model.

\paragraph{Eccentricity}
The steps for the Eccentricity NLI Entailment algorithm as described in \citet{lin2024eccentricity} are as follows, with more detailed explanations afterwards:
\begin{enumerate}
    \item Generate $M$ samples from the model.
    \item Use an NLI classifier to obtain predicted entailment scores between generations. Then, apply a softmax function to these logit values to obtain probabilities of entailment which are used as the similarity measure.
    \item Using the graph Laplacian and its corresponding embedding, calculate the average distance from the center as an eccentricity estimate. This serves as the input's uncertainty score.
\end{enumerate}
In step 2), the NLI classifier \citet{lin2024eccentricity} use is the DeBERTa-large model, which \citet{kuhn2023semantic} also use for Semantic Entropy. Instead of using the DeBERTa model to check for sequences entailing each other or not, the entailment logits are retrieved with a softmax applied afterwards to obtain a probability measure in $[0,1]$ of entailment as a proxy for similarity. 

In step 3), the embedding of a sequence can be represented through the eigenvectors of $L$, where $L$ is the symmetric normalized graph Laplacian. Specifically, if there are $M$ responses and $u_1,...,u_k \in \mathbb{R}^M$ are the $k$ eigenvectors of $L$ with the smallest eigenvalues, then the embedding $v_j$ of sequence $s_j$ is $[u_{1,j},...,u_{k,j}]$ \cite{luxburg07}. From this embedding, the eccentricity uncertainty is calculated by the following: 
\begin{equation}
    U(x) = \lvert\lvert[v_1'^\top,...,v_M'^\top]\lvert\lvert_{2}
\end{equation}
where $v_j' = v_j - \frac{1}{M}\sum^{M}_{j'=1}v_{j'}$ is the difference between $v_j$ and the average embedding.

\paragraph{LUQ}
Here, we discuss the calculation of similarity scores and the subsequent step to obtain the uncertainty measure in LUQ. As discussed in the paper, we are considering a simplified version of LUQ which does not include sentence splitting.
Formally, if the ``entailment'' value is $l_{entail}$ and the ``contradiction'' value is $l_{contradict}$: 
\begin{equation}
    similarity = \frac{e^{l_{entail}}}{e^{l_{entail}}+e^{l_{contradict}}}
\end{equation} Then, the uncertainty is computed via an average of the pairwise similarity scores, where higher similarity between responses indicates lower uncertainty. 

\section{Prompts}
\label{sec:appendix-prompts}

\begin{figure*}[]
\begin{tikzpicture}[node distance=0.5cm]
    \node[draw, rectangle, inner sep=5mm] (box1) {%
    \begin{varwidth}{0.95\textwidth}
      Please provide a single short argument \{``supporting" if support else ``attacking''\} the following claim. Construct the argument so it refers to the truthfulness of the claim. Only provide an argument if you think there is a valid and convincing \{``support" if support else ``attack''\} for this claim (there is a non-zero probability that the claim is \textbf{true}), otherwise return: N/A. \\
     Claim: \{statement\} \\
     Now take a deep breath and come up with an argument. \\
     Argument:
    \end{varwidth}
  };
  \node[draw, rectangle, inner sep=5mm, below=1cm of box1] (box2) {%
    \begin{varwidth}{0.95\textwidth}
      Please provide a single short argument \{``supporting" if support else ``attacking''\} the following claim. Construct the argument so it refers to the truthfulness of the claim. Only provide an argument if you think there is a valid and convincing \{``support'' if support else ``attack''\} for this claim (there is a non-zero probability that the claim is \textbf{\{``true'' if support else ``false''\}}), otherwise return: N/A. \\
     Claim: \{statement\} \\
     Now take a deep breath and come up with an argument. \\
     Argument:
    \end{varwidth}
  };
  \draw[->, thick] (box1.south) -- (box2.north);
\end{tikzpicture}
\caption{Prompt modification for the generation of supporting and attacking arguments, with the prompt from \citet{freedman2024} in the top box and the new prompt we use in the bottom box. The changed portion is shown in bold.}
\label{fig:prompts}
\end{figure*}
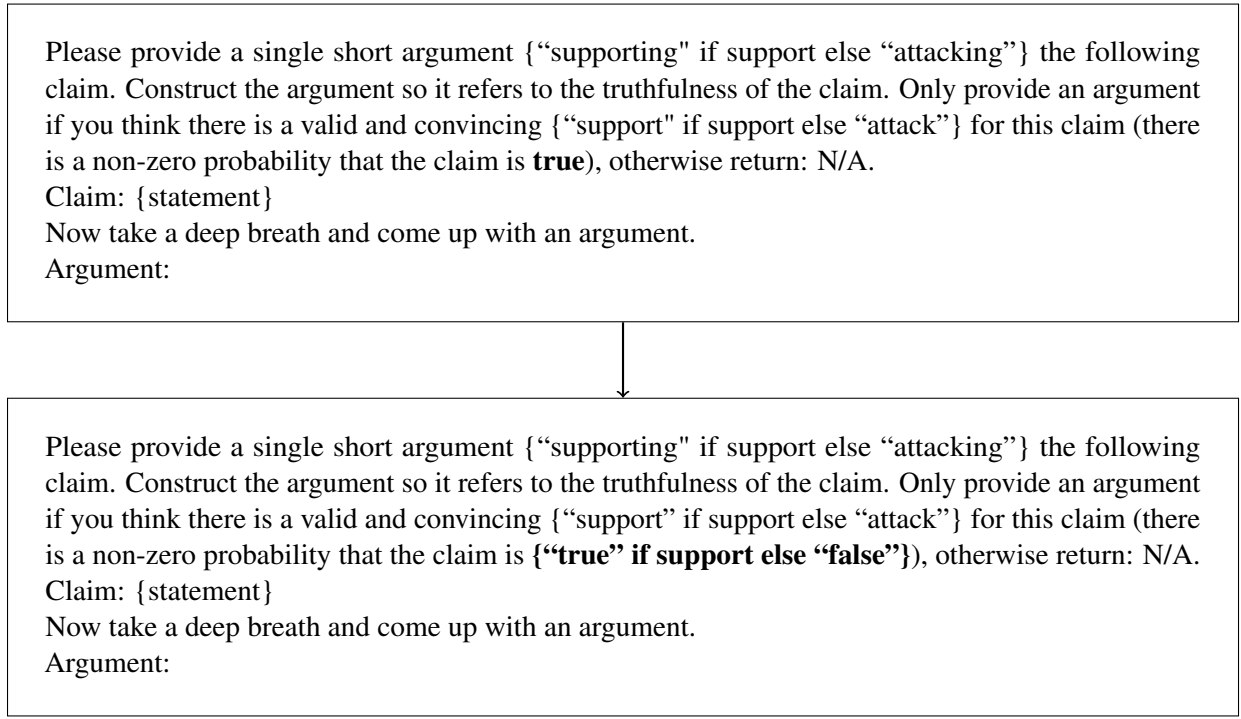

As mentioned in the paper, we employ the same prompts that \citet{freedman2024} use, except for the prompt used to generate the supporting and attacking arguments. For this prompt, we make a slight modification. In the original prompt, shown in the top box in Figure \ref{fig:prompts}, the section that tells the LLM to generate ``N/A'' if 
\begin{quote}
    ``there is a non zero probability that this claim is true''
\end{quote} does not distinguish between the supporting and attacking case. At times, this could lead to an ``N/A'' generation when prompted for an attacking argument with a topic claim that the LLM perceives to not be true, when in fact a strong attacking argument should be generated in this situation. Our new prompting accounts for this by conditioning this portion of the prompt on whether it is an attacking or supporting situation, as illustrated in Figure 
\ref{fig:prompts}. 

\begin{figure}[]
\includegraphics[width=\linewidth]{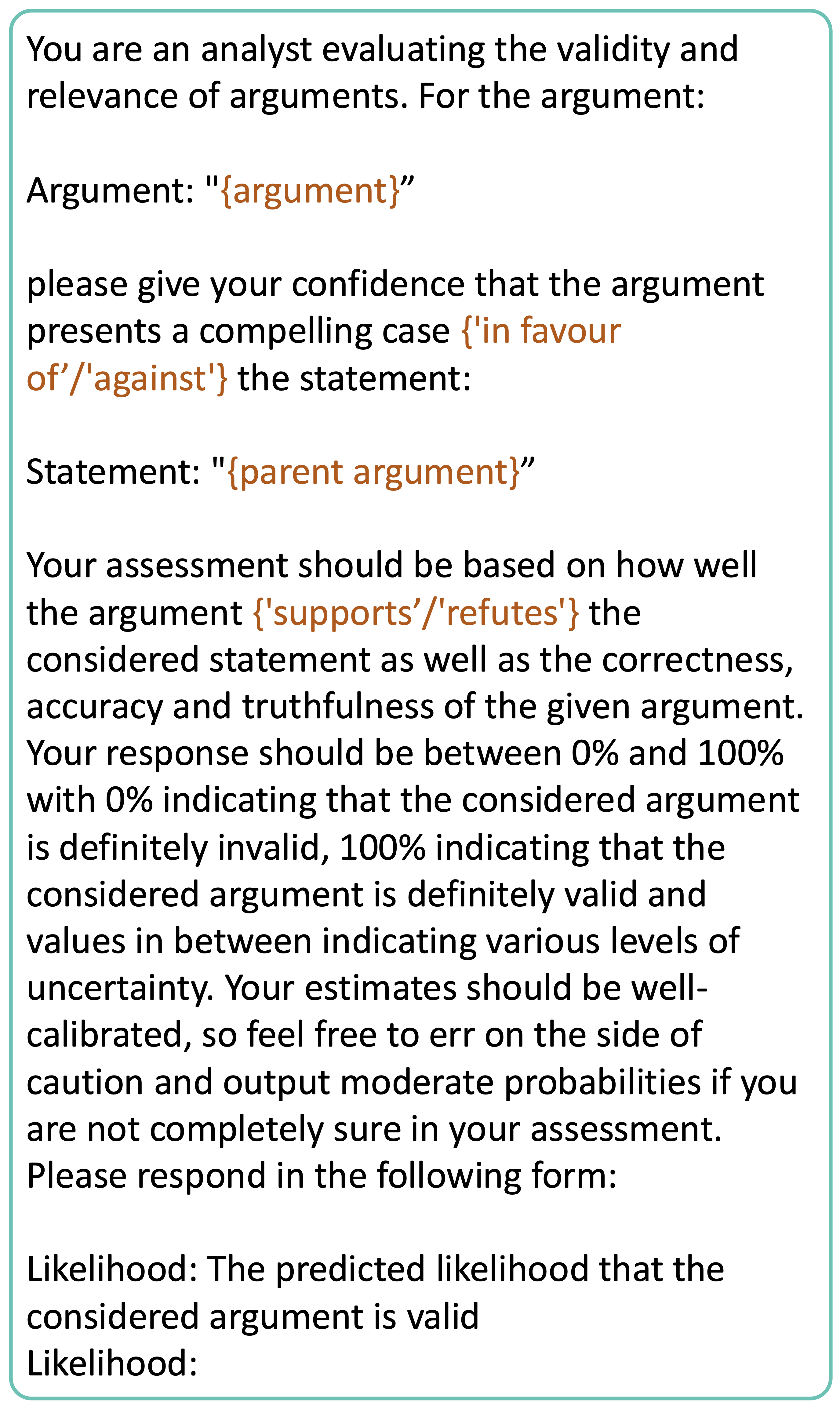}
\caption{The prompt used in the direct prompting method to obtain confidence scores for the generated supporting and attacking arguments (reproduced from \citet{freedman2024}).}
\label{fig:prompt-direct-args}
\end{figure}

For reference, we also provide the prompt used by \citet{freedman2024} and us to obtain confidence scores for the generated supporting and attacking arguments through direct prompting, in Figure \ref{fig:prompt-direct-args}.

\section{Datasets}
\label{sec:appendix-datasets}
The original QA datasets consist of question and answer pairs which are transformed into claims by an LLM and then manually checked and edited if needed to ensure faithfulness to the original data in \citet{freedman2024}. We use $500$ samples per experiment in order to have a sample size consistent with prior work on these datasets \cite{freedman2024} as well as provide a representative number of samples while keeping the experiments computationally tractable. 



\section{Experiment Parameters}
A key parameter for the LLMs in these experiments, especially considering the importance of semantic consistency in some of the UQ methods, is the temperature. We use a temperature of $0.7$ for the main LLM, which is used for direct prompting and the generation of supporting and attacking arguments. In addition, we use $p = 0.95$ top-p sampling and set the repetition penalty to $1.0$. For the generation of samples in the UQ procedures of Semantic Entropy, Eccentricity, and LUQ with LM-Polygraph, we use the default value of $1.0$ for the temperature, $p$, and repetition penalty.

\newpage
\newpage
\section{Results}
\label{sec:appendix-results}
\subsection{Brier Scores}
Table~\ref{tab:brier-scores} shows the table of Brier scores for all experiments. 

\begin{table*}[t]
    \centering
    \scriptsize
    \begin{tabular}{c|c|cccc|cccc|cccc}
        \multirow{3}{*}{\textbf{Model}} & 
        \multirow{3}{*}{\textbf{UQ Method}} & 
        \multicolumn{4}{c}{\textbf{TruthfulClaim}} &
        \multicolumn{4}{c}{\textbf{StrategyClaim}} &
        \multicolumn{4}{c}{\textbf{MedClaim}} \\
         &
         &
        \multicolumn{2}{c}{\textbf{D=1}} &  
        \multicolumn{2}{c}{\textbf{D=2}} &  
        \multicolumn{2}{c}{\textbf{D=1}} &  
        \multicolumn{2}{c}{\textbf{D=2}} &  
        \multicolumn{2}{c}{\textbf{D=1}} &  
        \multicolumn{2}{c}{\textbf{D=2}} \\
         &
         &
        \!\!\textbf{0.5 BS}\!\! &  
        \!\!\textbf{Est. BS}\!\! &
        \!\!\textbf{0.5 BS}\!\! &  
        \!\!\textbf{Est. BS}\!\! & 
        \!\!\textbf{0.5 BS}\!\! &  
        \!\!\textbf{Est. BS}\!\! &
        \!\!\textbf{0.5 BS}\!\! &  
        \!\!\textbf{Est. BS}\!\! & 
        \!\!\textbf{0.5 BS}\!\! &  
        \!\!\textbf{Est. BS}\!\! &
        \!\!\textbf{0.5 BS}\!\! &  
        \!\!\textbf{Est. BS}\!\! \\ 
        \hline
        \multirow{4}{*}{\makecell{Llama 3.1}} & 
        Direct Prompting & 
        0.217 & 0.253 & 0.217 & 0.244 & 0.239 & 0.324 & 0.252 & 0.324 & 0.243 & 0.274 & 0.251 & 0.272 \\
         & Semantic Entropy & 
        0.245 & 0.244 & 0.241 & 0.241 & 0.250 & 0.303 & 0.246 & 0.309 & 0.260 & 0.284 & 0.252 & 0.290 \\
         & Eccentricity & 
        0.292 & 0.282 & 0.263 & 0.257 & 0.282 & 0.338 & 0.264 & 0.331 & 0.310 & 0.342 & 0.271 & 0.314 \\
         & LUQ & 
        0.251 & 0.248 & 0.248 & 0.246 & 0.236 & 0.300 & 0.239 & 0.311 & 0.262 & 0.273 & 0.245 & 0.271 \\
        \hline
        \multirow{4}{*}{\makecell{Gemma 2}} & 
        Direct Prompting & 
        0.234 & 0.209 & 0.242 & 0.214 & 0.239 & 0.240 & 0.240 & 0.244 & 0.297 & 0.328 & 0.328 & 0.353 \\
         & Semantic Entropy & 
        0.263 & 0.196 & 0.247 & 0.191 & 0.274 & 0.242 & 0.259 & 0.231 & 0.265 & 0.305 & 0.244 & 0.300 \\
         & Eccentricity & 
        0.277 & 0.210 & 0.264 & 0.195 & 0.298 & 0.261 & 0.278 & 0.241 & 0.272 & 0.293 & 0.261 & 0.296 \\
         & LUQ & 
        0.250 & 0.207 & 0.252 & 0.199 & 0.265 & 0.241 & 0.254 & 0.237 & 0.257 & 0.296 & 0.254 & 0.298 \\
        \hline
        \multirow{4}{*}{\makecell{GPT-4o-mini}} & 
        Direct Prompting & 
        0.170 & 0.135 & 0.170 & 0.137 & 0.204 & 0.197 & 0.208 & 0.201 & 0.220 & 0.187 & 0.225 & 0.195 \\
         & Semantic Entropy & 
        N/A & N/A & N/A & N/A & N/A & N/A & N/A & N/A & N/A & N/A & N/A & N/A \\
         & Eccentricity & 
        0.274 & 0.187 & 0.267 & 0.169 & 0.269 & 0.220 & 0.253 & 0.202 & 0.270 & 0.223 & 0.258 & 0.200 \\
         & LUQ & 
        0.235 & 0.166 & 0.231 & 0.149 & 0.232 & 0.195 & 0.235 & 0.186 & 0.255 & 0.216 & 0.253 & 0.207 \\
        \hline
    \end{tabular}
    \caption{Brier scores ($\downarrow$) in all experiments. Semantic Entropy results for GPT-4o-mini are marked as N/A for the reasons discussed in Section~\ref{sec:models}. In the ``0.5 BS'' setting, the claim's base score is set at 0.5, while in ``Est. BS'' it is estimated through prompting.}
    \label{tab:brier-scores}
\end{table*}

\subsection{Confidence Intervals}
Table \ref{tab:truthfulCI}, Table \ref{tab:strategyCI}, and Table \ref{tab:medCI} show the confidence intervals from the bootstrapping procedure for the TruthfulClaim, StrategyClaim, and MedClaim experiments respectively. Confidence intervals where the values are either both negative or both positive indicate statistical significance. In these tables, if the values of the confidence interval are both negative, that means the first UQ method in the UQ Pair column performed statistically significantly worse than the second UQ method in this configuration, and if the values are both positive, then the first UQ method performed statistically significantly better than the second UQ method. 

\begin{table*}[t]
\centering
\scriptsize
\setlength{\tabcolsep}{4pt}
\renewcommand{\arraystretch}{0.95}
\begin{tabular}{c|c|cc|cc}
\textbf{Model} & \textbf{UQ Pair} &
\multicolumn{2}{c|}{\textbf{D=1}} &
\multicolumn{2}{c}{\textbf{D=2}} \\
& & \textbf{0.5 BS} & \textbf{Est. BS} & \textbf{0.5 BS} & \textbf{Est. BS} \\
\midrule
\multirow{6}{*}{\makecell{Llama 3.1}} &
Direct, SE  & (-0.0340, 0.0800) & (-0.0200, 0.0480) & (0.0040, 0.1160) & (-0.0240, 0.0380) \\
& Direct, Ecc & (0.0540, 0.1700) & (0.0140, 0.1020) & (0.0860, 0.2020) & (-0.0120, 0.0520) \\
& Direct, LUQ & (0.0160, 0.1240) & (-0.0380, 0.0320) & (0.0460, 0.1560) & (-0.0280, 0.0360) \\
& SE, Ecc     & (0.0320, 0.1480) & (0.0000, 0.0860) & (0.0260, 0.1440) & (-0.0160, 0.0380) \\
& SE, LUQ     & (-0.0120, 0.1060) & (-0.0520, 0.0180) & (-0.0180, 0.1020) & (-0.0280, 0.0220) \\
& Ecc, LUQ    & (-0.0980, 0.0140) & (-0.1060, -0.0160) & (-0.1000, 0.0140) & (-0.0440, 0.0120) \\
\midrule
\multirow{6}{*}{\makecell{Gemma 2}} &
Direct, SE  & (0.0940, 0.2180) & (-0.0460, 0.0180) & (0.0540, 0.1780) & (-0.0280, 0.0240) \\
& Direct, Ecc & (0.1080, 0.2300) & (-0.0200, 0.0560) & (0.1140, 0.2340) & (-0.0380, 0.0220) \\
& Direct, LUQ & (0.0580, 0.1680) & (-0.0120, 0.0500) & (0.0340, 0.1480) & (-0.0060, 0.0460) \\
& SE, Ecc     & (-0.0380, 0.0640) & (-0.0020, 0.0680) & (0.0100, 0.1060) & (-0.0321, 0.0220) \\
& SE, LUQ     & (-0.1020, 0.0160) & (0.0000, 0.0660) & (-0.0820, 0.0320) & (-0.0040, 0.0480) \\
& Ecc, LUQ    & (-0.1080, -0.0040) & (-0.0360, 0.0360) & (-0.1320, -0.0360) & (0.0000, 0.0560) \\
\midrule
\multirow{3}{*}{\makecell{GPT\mbox{-}4o\mbox{-}mini}} &
Direct, Ecc & (0.1840, 0.2880) & (0.0600, 0.1280) & (0.2140, 0.3200) & (0.0320, 0.0940) \\
& Direct, LUQ & (0.0840, 0.1920) & (0.0060, 0.0640) & (0.0920, 0.1980) & (0.0000, 0.0520) \\
& Ecc, LUQ    & (-0.1520, -0.0440) & (-0.0960, -0.0200) & (-0.1780, -0.0660) & (-0.0660, -0.0060) \\
\bottomrule
\end{tabular}
\caption{Confidence intervals for the accuracy differences between UQ methods in the TruthfulClaim experiments, with the order of the compared UQ methods given in the UQ Pair column (Direct = direct prompting, SE = Semantic Entropy, Ecc = Eccentricity).}
\label{tab:truthfulCI}
\end{table*}

\begin{table*}[t]
\centering
\scriptsize
\setlength{\tabcolsep}{4pt}
\renewcommand{\arraystretch}{0.95}
\begin{tabular}{c|c|cc|cc}
\textbf{Model} & \textbf{UQ Pair} &
\multicolumn{2}{c|}{\textbf{D=1}} &
\multicolumn{2}{c}{\textbf{D=2}} \\
& & \textbf{0.5 BS} & \textbf{Est. BS} & \textbf{0.5 BS} & \textbf{Est. BS} \\
\midrule
\multirow{6}{*}{\makecell{Llama 3.1}} &
Direct, SE  & (-0.0360, 0.0760) & (-0.0220, 0.0380) & (-0.0160, 0.0960) & (-0.0221, 0.0280) \\
& Direct, Ecc & (0.0080, 0.1160) & (-0.0040, 0.0720) & (0.0040, 0.1120) & (-0.0040, 0.0540) \\
& Direct, LUQ & (-0.0800, 0.0240) & (-0.0440, 0.0160) & (-0.0780, 0.0340) & (-0.0360, 0.0160) \\
& SE, Ecc     & (-0.0100, 0.0960) & (-0.0100, 0.0620) & (-0.0360, 0.0720) & (-0.0020, 0.0460) \\
& SE, LUQ     & (-0.1060, 0.0100) & (-0.0540, 0.0080) & (-0.1200, -0.0020) & (-0.0340, 0.0100) \\
& Ecc, LUQ    & (-0.1440, -0.0400) & (-0.0860, -0.0100) & (-0.1340, -0.0260) & (-0.0600, -0.0080) \\
\midrule
\multirow{6}{*}{\makecell{Gemma 2}} &
Direct, SE  & (0.1160, 0.2420) & (0.0040, 0.0780) & (0.1020, 0.2220) & (-0.0220, 0.0340) \\
& Direct, Ecc & (0.1200, 0.2440) & (0.0280, 0.1160) & (0.1420, 0.2620) & (-0.0100, 0.0620) \\
& Direct, LUQ & (0.0620, 0.1780) & (-0.0020, 0.0700) & (0.0580, 0.1720) & (-0.0120, 0.0440) \\
& SE, Ecc     & (-0.0440, 0.0480) & (-0.0020, 0.0660) & (-0.0100, 0.0880) & (-0.0080, 0.0500) \\
& SE, LUQ     & (-0.1160, -0.0040) & (-0.0440, 0.0320) & (-0.1020, 0.0080) & (-0.0160, 0.0360) \\
& Ecc, LUQ    & (-0.1140, -0.0100) & (-0.0741, -0.0020) & (-0.1380, -0.0360) & (-0.0400, 0.0200) \\
\midrule
\multirow{3}{*}{\makecell{GPT\mbox{-}4o\mbox{-}mini}} &
Direct, Ecc & (0.0480, 0.1480) & (0.0300, 0.0940) & (0.1020, 0.2100) & (-0.0180, 0.0420) \\
& Direct, LUQ & (-0.0160, 0.0900) & (-0.0100, 0.0500) & (0.0080, 0.1080) & (-0.0340, 0.0200) \\
& Ecc, LUQ    & (-0.1160, -0.0060) & (-0.0760, -0.0080) & (-0.1520, -0.0440) & (-0.0480, 0.0140) \\
\bottomrule
\end{tabular}
\caption{Confidence intervals for the accuracy differences between UQ methods in the StrategyClaim experiments, with the order of the compared UQ methods given in the UQ Pair column (Direct = direct prompting, SE = Semantic Entropy, Ecc = Eccentricity).}
\label{tab:strategyCI}
\end{table*}

\begin{table*}[t]
\centering
\scriptsize
\setlength{\tabcolsep}{4pt}
\renewcommand{\arraystretch}{0.95}
\begin{tabular}{c|c|cc|cc}

\textbf{Model} & \textbf{UQ Pair} &
\multicolumn{2}{c|}{\textbf{D=1}} &
\multicolumn{2}{c}{\textbf{D=2}} \\
& & \textbf{0.5 BS} & \textbf{Est. BS} & \textbf{0.5 BS} & \textbf{Est. BS} \\
\midrule
\multirow{6}{*}{\makecell{Llama 3.1}} &
Direct, SE  & (-0.0080, 0.1020) & (-0.0360, 0.0540) & (-0.0340, 0.0700) & (-0.0060, 0.0800) \\
& Direct, Ecc & (0.0519, 0.1700) & (0.0360, 0.1400) & (0.0240, 0.1460) & (0.0100, 0.1020) \\
& Direct, LUQ & (-0.0400, 0.0720) & (-0.0360, 0.0600) & (-0.0520, 0.0540) & (-0.0500, 0.0420) \\
& SE, Ecc     & (0.0060, 0.1200) & (0.0320, 0.1240) & (0.0080, 0.1220) & (-0.0100, 0.0480) \\
& SE, LUQ     & (-0.0840, 0.0220) & (-0.0460, 0.0500) & (-0.0720, 0.0360) & (-0.0740, -0.0100) \\
& Ecc, LUQ    & (-0.1560, -0.0320) & (-0.1240, -0.0280) & (-0.1440, -0.0240) & (-0.0920, -0.0280) \\
\midrule
\multirow{6}{*}{\makecell{Gemma 2}} &
Direct, SE  & (-0.0100, 0.1140) & (-0.0120, 0.0520) & (-0.0300, 0.0880) & (-0.0180, 0.0260) \\
& Direct, Ecc & (-0.0260, 0.0980) & (-0.0200, 0.0580) & (0.0140, 0.1420) & (-0.0280, 0.0300) \\
& Direct, LUQ & (-0.0500, 0.0580) & (-0.0160, 0.0380) & (-0.0220, 0.0880) & (-0.0260, 0.0220) \\
& SE, Ecc     & (-0.0720, 0.0420) & (-0.0420, 0.0420) & (-0.0040, 0.1060) & (-0.0300, 0.0240) \\
& SE, LUQ     & (-0.1060, 0.0120) & (-0.0440, 0.0240) & (-0.0540, 0.0620) & (-0.0320, 0.0200) \\
& Ecc, LUQ    & (-0.0840, 0.0220) & (-0.0481, 0.0280) & (-0.0940, 0.0020) & (-0.0280, 0.0240) \\
\midrule
\multirow{3}{*}{\makecell{GPT\mbox{-}4o\mbox{-}mini}} &
Direct, Ecc & (0.0560, 0.1620) & (0.0280, 0.0960) & (0.0680, 0.1880) & (-0.0060, 0.0540) \\
& Direct, LUQ & (0.0320, 0.1500) & (0.0220, 0.0900) & (0.0740, 0.1920) & (-0.0180, 0.0320) \\
& Ecc, LUQ    & (-0.0760, 0.0360) & (-0.0460, 0.0340) & (-0.0500, 0.0600) & (-0.0520, 0.0160) \\
\bottomrule
\end{tabular}
\caption{Confidence intervals for the accuracy differences between UQ methods in the MedClaim experiments, with the order of the compared UQ methods given in the UQ Pair column (Direct = direct prompting, SE = Semantic Entropy, Ecc = Eccentricity).}
\label{tab:medCI}
\end{table*}

\end{document}